\documentclass[10pt, a4paper]{article}

\usepackage[]{lrec-coling2024} 

\usepackage{caption}
\newcommand{\hide}[1]{}

\newcommand{\DAD}{{\v{D}}} 

\newcommand{\SHADDA}{{$\sim$}}

\usepackage{arabtex}
\usepackage{utf8}
\setcode{utf8}
\vocalize

\title{The SAMER Arabic Text Simplification Corpus}

\name{Bashar Alhafni, Reem Hazim, Juan Piñeros Liberato,\\
\textbf{\large{Muhamed Al Khalil, Nizar Habash}} } 

\address{Computational Approaches to Modeling Language (CAMeL) Lab \\
         New York University Abu Dhabi \\
         \{alhafni,rh3015,juanpl,muhamed.alkhalil,nizar.habash\}@nyu.edu\\}

\abstract{
We present the \textbf{SAMER~Corpus}, the first manually annotated Arabic parallel corpus for text simplification targeting school-aged learners. Our corpus comprises texts of 159K words selected from 15 publicly available Arabic fiction novels most of which were published between 1865 and 1955. Our corpus includes readability level annotations at both the document and word levels, as well as two simplified parallel versions for each text targeting learners at two different readability levels. We describe the corpus selection process, and outline the guidelines we followed to create the annotations and ensure their quality. Our corpus is publicly available to support and encourage research on Arabic text simplification, Arabic automatic readability assessment, and the development of Arabic pedagogical language technologies.
\\ \newline  }

\begin{document}
\maketitleabstract


\section{Introduction}

Text simplification aims to reduce the complexity of a text while maintaining the overall grammaticality and core content. This is achieved through a series of different rewriting transformations at both the lexical and syntactic levels. Having simplified versions of texts has many benefits to users with cognitive and reading disorders \cite{carroll:1998:practical,rello:2013:frequent,evans-etal-2014-evaluation}, second language learners \cite{paetzold-specia-2016-understanding}, and native speakers with low literacy levels \cite{candido-etal-2009-supporting,watanbe:2009:facilita}.  Text simplification can also be used as a preprocessing step to improve performance on other downstream NLP tasks such as machine translation \cite{stajner-popovic-2016-text,hasler:2017:mt} and summarization \cite{silveira:2012:using}. Simplifying text can be achieved in multiple ways, depending on the target audience: for example, second-language learners and school-aged learners might struggle with texts containing  different vocabulary items that go beyond their \textit{respective} language proficiency levels. Yet, research in text simplification has mostly focused on developing models that produce a single simplification for a given input without the possibility of adapting to different users' needs. Moreover, studies on text simplification are heavily focused on English due to the availability of large parallel simplification corpora \cite{alva-manchego-etal-2020-data}.
For other languages, data is limited in terms of size and domain. And when it comes to morphologically rich languages, particularly Arabic, we are not aware of any manually annotated publicly available datasets for text simplification.  

\begin{figure*}[t!]
    \centering
    \includegraphics[width=1.0\textwidth]{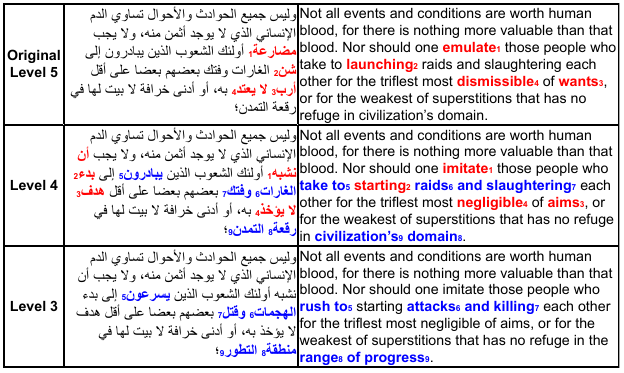}
    \captionof{table}{An example consisting of two sentences (in three punctuated fragments) and its simplified parallels from the Arabic novel ``The Forest of Truth'' \cite{marrash:1865}. 
Level~4 is the simplified version of the original text where all level 5 words (in \textcolor{red}{\textbf{red}}) are simplified to level 4 or lower according to \newcite{AlKhalil:2018:leveled}'s readability lexicon. Level~3 is the simplified version of the Level~4 text where all level 4 words (in \textcolor{blue}{\textbf{blue}}) are simplified to level 3 or lower. The words that change during the simplification are co-indexed with subscript numbers and carry the same color marking of the level they changed from. 
   }
    \label{tab:intro-example}
\end{figure*}

In this paper, we present the \textbf{SAMER Corpus}, the first manually annotated Arabic parallel corpus for text simplification targeting school-aged learners. Our corpus comprises texts of 159K words selected from 15 publicly available Arabic fiction novels, 14 of which were published between 1865 and 1955, and one famous philosophical novel written in the 12th century. We focus on lexical simplification, i.e., replacing complex words in a given text with simpler alternatives of equivalent meaning. We define the text simplification task as paraphrasing into a controlled language with a vocabulary that is anchored in a readability-leveled lexicon. Our corpus includes readability level annotations at both the document and word levels, as well as two simplified parallels for each text, targeting school-aged learners at two different readability levels. We describe the corpus selection process and outline the guidelines we followed to create the annotations and ensure their quality. Our corpus is publicly available to support and encourage research on Arabic text simplification and automatic readability assessment, as well as the development of pedagogical language technologies. Table~\ref{tab:intro-example} presents an example from our simplification corpus. 

This work is one of the publicly available resources created by the \textbf{\textit{Simplification  of Arabic Masterpieces for Extensive Reading} (SAMER)} project \cite{AlKhalil:2017:simplification},\footnote{\url{http://samer.camel-lab.com/}} which includes a readability leveled lexicon \cite{AlKhalil:2018:leveled,jiang-etal-2020-online}, and a Google Doc add-on \cite{hazim-etal-2022-arabic}.

Next, we discuss related work and basic Arabic linguistic facts. In Section~\ref{corpus}, we introduce our corpus, and describe its selection and annotation process. Section~\ref{overview} presents the corpus statistics and discusses its simplification patterns.

\section{Related Work}

We first provide an overview of existing work related to general text simplification approaches and datasets, before zooming in on Arabic text simplification specifically.

\subsection{Text Simplification Approaches} When simplifying text, different rewriting transformations are performed. Such transformations range from \textit{lexical simplification}, which is the process of replacing complex words or phrases with simpler synonyms, to \textit{syntactic simplification}, which includes splitting or reordering sentences. Most research on text simplification has focused on simplifying individual sentences. This allows for easier curation of data and reduces the complexity of modeling. Several modeling approaches for text simplification have been explored. This includes syntactic simplification, lexical simplification, and end-to-end models that can learn and induce both syntactic and lexical transformations. 

Efforts on lexical simplification often involve four subtasks: Complex Word Identification, Substitution Generation, Substitution Selection, and Substitution Ranking \cite{shardlow-2014-open} with approaches ranging from lexicon-based lookups \cite{elhadad-sutaria-2007-mining,kajiwara-etal-2013-selecting} to statistical machine learning systems \cite{paetzold-specia-2016-semeval,gooding-kochmar-2018-camb,gooding-kochmar-2019-recursive}, and more recently, deep learning models \cite{de-hertog-tack-2018-deep,maddela-xu-2018-word,qiang:2020:lexical,qiang:2021:lsbert,sheang-etal-2022-controllable}.

In contrast, efforts for syntactic simplification focused on rule-based systems \cite{chandrasekar-etal-1996-motivations,Gasperin2009NaturalLP} and statistical machine learning techniques by drawing inspirations from phrase- and tree-based statistical machine translation models \cite{specia:2010:mt,zhu-etal-2010-monolingual,wubben-etal-2012-sentence}.  

Finally, end-to-end text simplification approaches are the dominant paradigm in the literature. End-to-end models can perform multiple simplification transformations simultaneously, while learning very specific and complex rewriting patterns. The majority of approaches treat text simplification as a monolingual machine translation task, where both statistical \cite{coster-kauchak-2011-learning,wubben-etal-2012-sentence,xu-etal-2016-optimizing} and neural machine translation models \cite{nisioi-etal-2017-exploring,zhang-lapata-2017-sentence,stajner-nisioi-2018-detailed,vu-etal-2018-sentence,guo-etal-2018-dynamic,zhao-etal-2018-integrating,surya-etal-2019-unsupervised,martin-etal-2020-controllable,maddela-etal-2021-controllable} were explored. These models require large amounts of parallel training data and provide little control or adaptability to different aspects of simplification, which inhibits interpretability and explainability. Moreover, these models are typically slow as they employ autoregressive decoders, i.e., output texts are generated in a sequential, non-parallel fashion. To address some of these limitations, sequence labeling and edit-based models were explored \cite{alva-manchego-etal-2017-learning,omelianchuk-etal-2021-text}.

\subsection{Text Simplification Datasets} Most of the recent advancements in text simplification have focused on English, which is attributed to the availability of large parallel datasets. Most of the English datasets \cite{zhu-etal-2010-monolingual,coster-kauchak-2011-simple,woodsend:2011,kauchak-2013-improving,hwang-etal-2015-aligning,kajiwara-komachi-2016-building,zhang-lapata-2017-sentence} were created by automatically aligning sentences from English Wikipedia and Simple English Wikipedia, a simplified version of English Wikipedia that is primarily aimed at English learners, but which can also be beneficial for students, children, and adults with learning difficulties. 

Although the large scale and availability of Wikipedia-based corpora is a strong asset to build simplification models,  studies have shown that Wikipedia-based text simplification corpora are limited in various ways, including the presence of noisy instances caused by misalignments and a lack of variety in simplification transformations \cite{yasseri:2012}. To address these limitations, several manually annotated datasets were introduced such as the Newsela Corpus \citelanguageresource{xu-etal-2015-problems}, TurkCorpus \cite{xu-etal-2016-optimizing}, HSplit \citelanguageresource{sulem-etal-2018-simple}, SimPA \cite{scarton-etal-2018-simpa}, and OneStopEnglish \citelanguageresource{vajjala-lucic-2018-onestopenglish}.

While the most popular (and generally larger) resources available for simplification are in English, there are some resources that have been built for other languages such as Basque \citelanguageresource{gonzalez-dios-etal-2014-simple}, Brazilian-Portuguese \citelanguageresource{caselie:2009,aluisio-gasperin-2010-fostering}, Danish \citelanguageresource{klerke-sogaard-2012-dsim}, French \citelanguageresource{grabar-cardon-2018-clear,gala-etal-2020-alector,cardon-grabar-2020-french}, German \citelanguageresource{klaper-etal-2013-building,battisti-etal-2020-corpus,aumiller-gertz-2022-klexikon}, Italian \citelanguageresource{brunato-etal-2015-design,tonelli:2016,miliani-etal-2022-neural}, Japanese \citelanguageresource{goto-etal-2015-japanese,hayakawa-etal-2022-jades}, Spanish \citelanguageresource{bott-etal-2012-text,xu-etal-2015-problems,saggion:2015}, Russian \citelanguageresource{dmitrieva-tiedemann-2021-creating,sakhovskiy:2021}, and Urdu \citelanguageresource{qasmi-etal-2020-simplifyur}.

\begin{table}[t]
\centering
    \includegraphics[width=0.96\columnwidth]{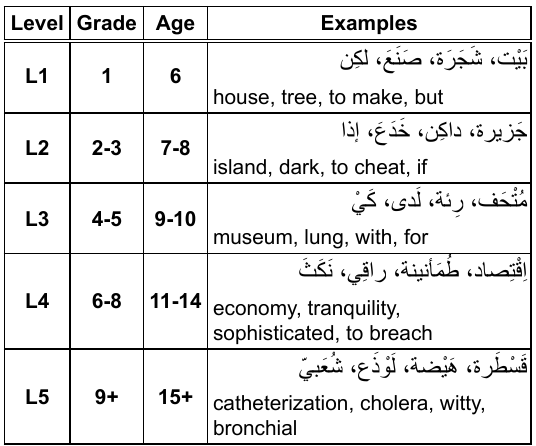}
%
\caption{The five readability levels, their grade equivalencies, and lemma and English gloss examples, abridged from \newcite{al-khalil-etal-2020-large}. }
\label{tab:readabilitylevels}
\end{table}

\subsection{Arabic Text Simplification} 
While there are some limited efforts to publishing simplified and abridged texts in Arabic, the only simplification resource that was used in NLP, to our knowledge, is the simplified version of ``Saaq al-Bambuu (The Bamboo Stalk)'' \cite{Sanousi:2012}, an internationally acclaimed Arabic novel, that has been rewritten for Arabic-as-a-second-language learners \citelanguageresource{familiar2017saud}. \newcite{khallaf-etal-2022-towards} automatically aligned sentences from ``Saaq al-Bambuu'' and sampled 2,980 parallel sentences from the original and simplified books at two different literacy levels. Unfortunately, due to copyright restrictions, the corpus is not publicly available. 

More recently, there have been grassroots efforts to create Arabic text simplification resources \cite{AlKhalil:2017:simplification,AlKhalil:2018:leveled,al-khalil-etal-2020-large,jiang-etal-2020-online,hazim-etal-2022-arabic}. As part of the SAMER Project, \newcite{al-khalil-etal-2020-large} developed a 26K-lemma lexicon with a five-level readability scale, later extended to 40K lemmas \cite{jiang-etal-2020-online}. 
The levels range from \textbf{L1} (Low Difficulty/Easy Readability) to \textbf{L5} (High Difficulty/Hard Readability). See examples in Table~\ref{tab:readabilitylevels}. 
%
We use this lexicon as our main reference for readability leveling when creating our corpus to ensure that our simplified texts are appropriate for the audience we are targeting. \newcite{hazim-etal-2022-arabic} created a Google Docs add-on for automatic Arabic word-level readability visualization, which includes a lemmatization component that is connected to the five-level readability lexicon \cite{al-khalil-etal-2020-large} and Arabic WordNet-based substitution suggestions \cite{Black:2006:introducing}. The add-on enables users to edit texts easily based on a specific target readability level. We use the add-on as our main annotation tool to enable human annotators to identify text readability levels and to simplify texts in a controlled setting.

\section{Arabic Linguistic Facts}
Arabic is a morphologically rich language that inflects for gender, number, person, case, state, aspect, mood and voice, in addition to numerous attachable clitics such as prepositions, particles, and pronouns \cite{Habash:2010:introduction}. This results in a large number of forms for any particular word, with different morpho-syntactic restrictions. In addition to its morphological richness, Arabic is orthographically ambiguous and uses diacritics to specify short vowels and consonantal doubling. These diacritics are optional and often omitted, leaving readers to decipher words using contextual and templatic morphology clues. 
Orthographic ambiguity and morphological richness interact heavily with each other. For instance, the word \<درسها> \textit{drshA} has different readings with varying analyses including \<دَرَّسَها> \textit{dar{\SHADDA}asa+hA} `he taught her', \<دَرَسَها> \textit{darasa+hA} `he studied it', and \<دَرْسُها> \textit{darsu+hA} `her lesson'. Moreover, these different readings have three unique lemmas (lexical entries) that abstract away from the various inflections: \<دَرَّس> \textit{dar{\SHADDA}as
`taught'}, \<دَرَس> \textit{daras `studied'}, and \<دَرْس> \textit{dars `lesson'}.  This issue highlights the complexity of lexical simplification in Arabic, which cannot be accomplished through a simple word dictionary lookup.

The SAMER project lexicon \cite{al-khalil-etal-2020-large} discussed above anchors readability at the lemma representation of the words. 
There are publicly available Arabic morphological disambiguation and lemmatization tools that diacritize Arabic text and map each word to a predicted lemma \cite{pasha-etal-2014-madamira,obeid-etal-2022-camelira}. 

In this paper we focus on Modern Standard Arabic (MSA) and do not discuss dialectal Arabic variants, which are not typically used in high literature.


\begin{figure*}[th!]
    \centering
    \includegraphics[width=0.96\textwidth]{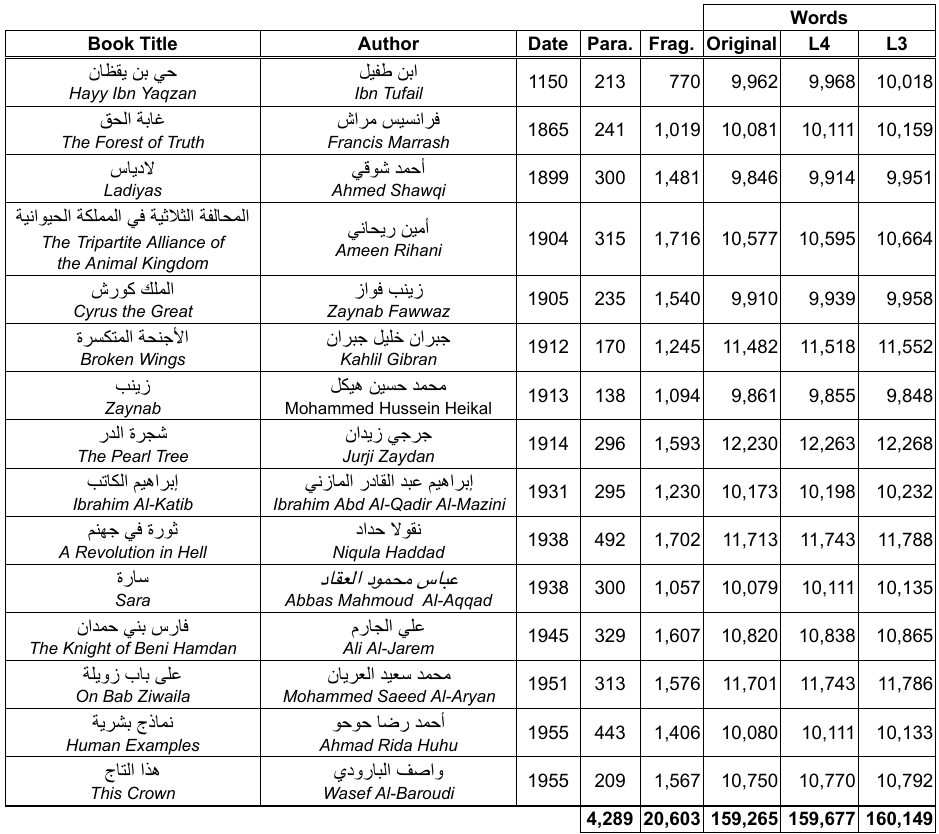}
    \captionof{table}{
    The 15 books we selected to create the SAMER Arabic Text Simplification 
    Corpus. L4-Words and L3-Words refer to the number of words in the Level~4 and Level~3 simplified versions of the books, respectively. Para. is \textit{Paragraphs}; and Frag. is \textit{Punctuated Fragments}.
    }
    \label{tab:simplification-books}
\end{figure*}

\section{The SAMER Arabic Text Simplification Corpus}
\label{corpus} 
\subsection{Corpus Selection}
In making the specific selection of texts to annotate, we aimed to cover Arabic fiction novels from a large historical span with high readability levels  (i.e., hard to read) targeted toward proficient Arabic readers. But most importantly, we wanted the texts to be publicly available (out of copyright or under open licenses). We identified and selected 15 Arabic fiction novels that match these requirements from the online catalog of the Hindawi Foundation.\footnote{\url{http://www.hindawi.org/}} Most of the novels were published between 1865 and 1955 and one philosophical novel was from the 12th century.  From each novel, we extracted the first $\sim$10K words based on chapter boundaries, and we ended up with $\sim$159K words in total. To make the annotation task easier, we further segmented the chapters based on paragraph boundaries if they consisted of more than 1,500 words. This resulted in 4,289 paragraphs. We were restricted by an annotation budget that affected how many novels we could work with. There were many interesting options that we decided to leave to future annotation follow-up projects. Table~\ref{tab:simplification-books} presents the list of the  selected books. 

\begin{table*}[t!]
    \centering
        \begin{tabular}{c c}
        (a) & \includegraphics[width=0.6\textwidth]{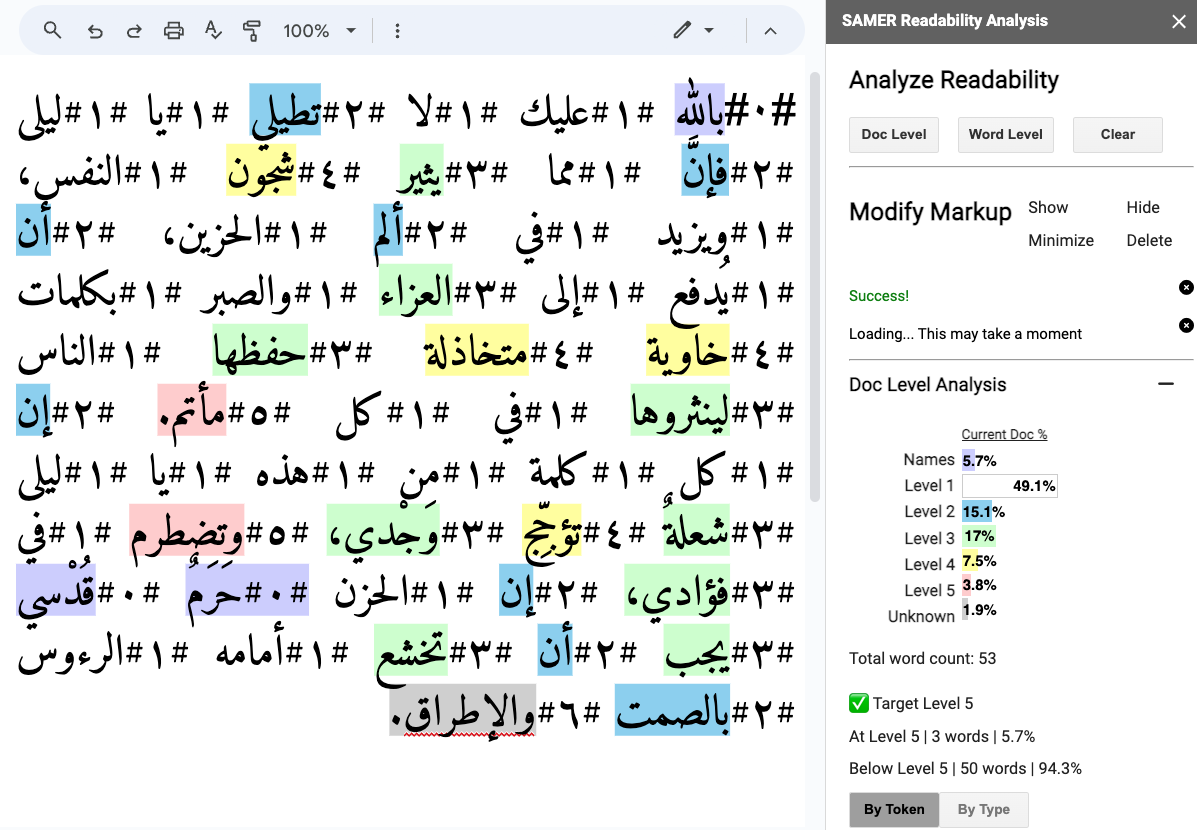} \\\hline\hline
        (b) & \includegraphics[width=0.6\textwidth]{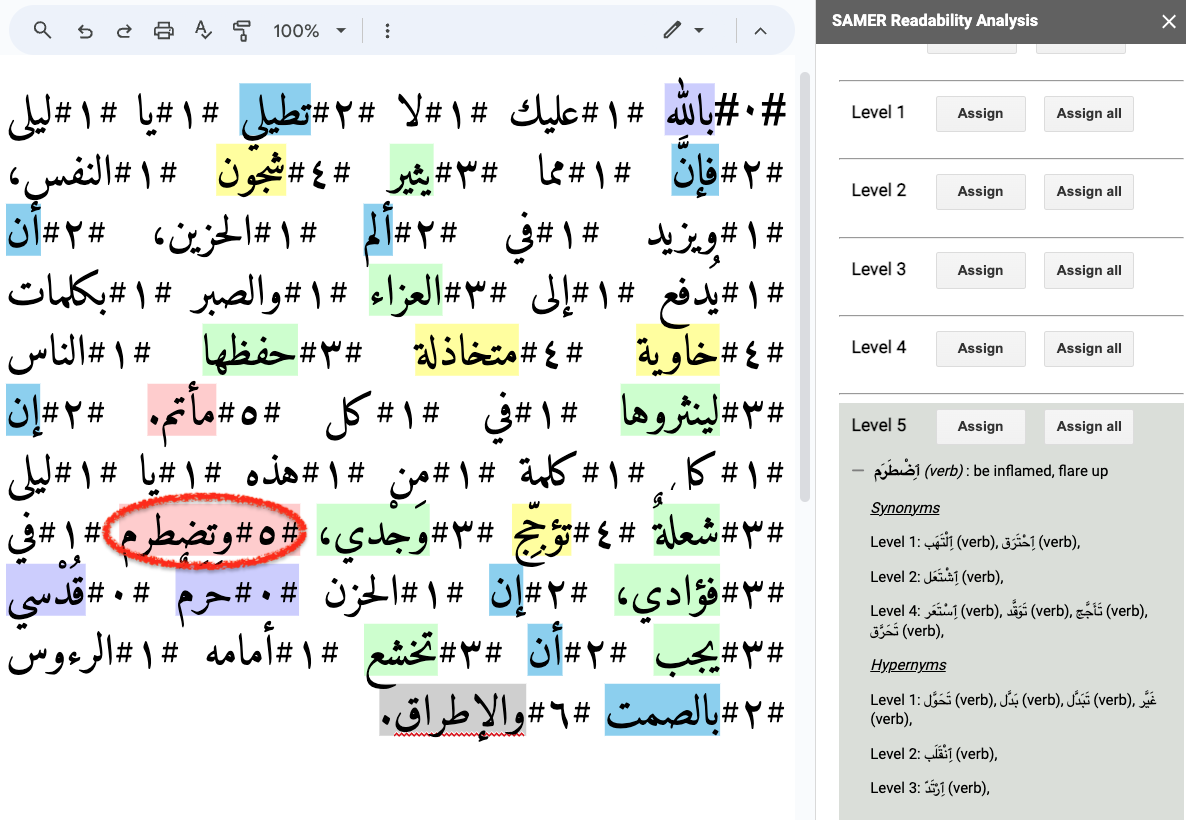}  \\\hline\hline
     \end{tabular}
    \captionof{figure}{The Google Doc add-on annotation interface introduced by \newcite{hazim-etal-2022-arabic}. Figure (a) is a visualization of word-level and document-level readability. Figure (b) is an example of selecting a specific word to identify all of its analyses and their readability levels. The add-on has multiple markup viewing modes. We show the maximally explicit view where each word is prefixed by an Indo-Arabic digit indicating its level.}
    \label{fig:interface}
\end{table*}

\hide{
\begin{table*}[t!]
    \centering
        \begin{tabular}{p{0.47\textwidth} || p{0.47\textwidth}}
            \multicolumn{1}{c}{(a)} & \multicolumn{1}{c}{(b)} \\
             \includegraphics[width=0.47\textwidth]{annotation_interface.png} &
             \includegraphics[width=0.47\textwidth]{annotation_interface_2.png}  \\
                
            \end{tabular}
    \captionof{figure}{The Google Doc add-on annotation interface introduced by \newcite{hazim-etal-2022-arabic}. Figure (a) is a visualization of word-level and document-level readability. Figure (b) is an example of selecting a specific word to identify all of its analyses and their readability levels.}
    \label{fig:interface}
\end{table*}
}
\subsection{Corpus Annotation}
Our goal is to simplify the Arabic fiction novels we selected so that they can be targeted toward school-age learners. Given Arabic morphological richness, we consider the lexical and syntactic aspects of text simplification to be independent and complementary to each other. In our work, we focus on lexical simplification for a number of reasons. First, research has shown that lexical simplification improves text readability \cite{leroy:2012}, benefiting those with lower literacy levels and non-expert readers \citelanguageresource{xu-etal-2015-problems}. Second, we want to create our dataset in a controlled way to avoid context inconsistencies that may result from changing the syntactic structure of Arabic text. This process ensures that the simplified text is indeed of a lower complexity while being semantically equivalent to the original, and grammatically correct.


To ensure that the simplified texts are indeed appropriate for our target audience throughout the annotation process, we use the five readability levels that were defined by \newcite{al-khalil-etal-2020-large} and exemplified in Table~\ref{tab:readabilitylevels}.  For the purpose of the corpus annotation, we consider the document readability level to be equal to the highest readability level found among the words in the document. Our focus is on the needed competence level to easily read a document rather than the use of documents as language-learning artifacts. Based on this and given the nature of the documents we selected, all documents will have a readability of \textbf{Level~5} (although some sentences in them may be of lower/easier readability levels). We focus on producing two simplified versions for each document targeting \textbf{Level~4} (grades 6--8) and \textbf{Level~3} (grades 4-5), respectively.

\subsection{Annotation Interface}
Three professional female computational linguists, all of whom are native speakers of Arabic, were hired through a linguistic annotation firm to complete the task.\footnote{\url{https://www.ramitechs.com/}} The annotators were provided with 146 Google Docs that included the chapters that needed to be simplified. Every Google Doc was equipped with the add-on developed by \newcite{hazim-etal-2022-arabic} which served as the primary annotation interface. Before sharing the documents with the annotators, all the documents were automatically labeled with their word-level readability using a Python API version of the add-on developed by \newcite{hazim-etal-2022-arabic}. The Python API performs the same functionality as the add-on, but it relies on CAMeL Tools \cite{obeid-etal-2020-camel} to tokenize and disambiguate the words in context for each chapter. Specifically, the API leverages the BERT unfactored morphological disambiguator developed by \newcite{inoue2022morphosyntactic} to retrieve the most probable lemma and part-of-speech (POS) tag for each word. After that, the lemmas and POS tags are looked up in the lemma-based readability lexicon developed by \newcite{al-khalil-etal-2020-large} to identify the readability levels of the words. The add-on employs two additional levels to classify proper nouns (Level 0) and unknown words (Level 6) that are not present in the lexicon. After the word-level readability labeling is done, each chapter is loaded into a Google Doc where the add-on highlights words with different colors according to their readability levels.
Figure~\ref{fig:interface}(a) presents a visualization of using the add-on to analyze the readability of a short segment of an Arabic novel.
The interface provides a summary of the text's readability distribution levels in a bar chart colored consistently with the readability level word highlights. 
 The interface also provides the option for explicit word-level readability markup by adding a prefix (\textit{\#<level>\#}) in front of each word, where \textit{<level>} is an Indo-Arabic digit indicating the word readability level (see Figure~\ref{fig:interface}).
Moreover, the add-on incorporates the Arabic WordNet \cite{Black:2006:introducing} and supports word substitution by displaying suggestions for related words and phrases, e.g., synonyms, antonyms, hypernyms, and hyponyms, with different readability levels. Figure~\ref{fig:interface}(b) shows the result of selecting a specific word (\<وتضطرم> {\it wt{\DAD}Trm} `be inflamed'). A sidebar appears showing the different lemma analyses by their readability levels.  If the annotators decide to change the automatically assigned readability level, they can either change it directly manually, or by clicking on the \texttt{Assign} button to change that specific word's readability level markup or the \texttt{Assign All} button to change all of its occurrences in the document.

\begin{figure*}[t!]
    \centering
    \includegraphics[]{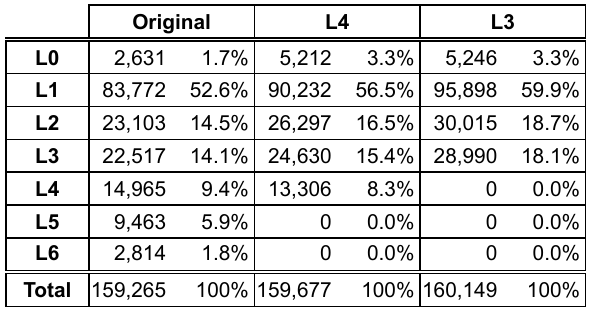}
    \captionof{table}{Readability levels statistics of the words in our corpus over the original text (Original), Level~4 simplified text (L4), and Level~3 simplified text (L3).
   }
    \label{tab:level-stats}
\end{figure*}

\subsection{Annotation Guidelines}
 The annotation process starts with determining the original text readability level. To do so, the annotators are instructed to check the word-level readability levels assigned to each word using the add-on. The annotators would then determine the readability level of the document based on the highest readability level found among the words in the document. Since the readability levels are automatically generated, there might be cases where the predicted levels do not reflect the true readability of the words. This could happen either because the lemmas of some words are not in the lexicon or due to morphological tagging errors when the lemmas and POS tags are identified. Therefore, the annotators are asked to use the add-on to fix all of the anomalies related to the readability levels of the words. In cases where correct readability levels are not among the add-on suggestions, the annotators are asked to adjust the levels manually.
 
 After making the corrections, the annotators need to re-run the add-on analysis to determine the document's readability level. If the document has a readability of \textbf{Level~5}, then the annotators have to first simplify it to \textbf{Level~4}, and then to \textbf{Level~3} starting from \textbf{Level~4}. If the original document has a readability \textbf{Level~4} then the annotators need to simplify it to \textbf{Level~3}. However, if the document has a readability \textbf{Level~3} then no simplification is needed. When simplifying text, the annotators are allowed to perform minimal replacements, deletions, and insertions that are needed to reduce the readability level of the document. Each of these operations might involve more than one word at a time. However, the simplification should be done carefully so that the original meaning of the text is preserved, and as such no abridgement or summarization should take place. Once the simplification is done, the annotators need to re-run the add-on to analyze the text and verify that the document readability level has been adjusted as intended. At the end of the annotation process, each document will have \textbf{three} parallel versions: (1)~The \textbf{Original} text; (2)~\textbf{Level~4} simplified text; (3)~and \textbf{Level~3} simplified text. Each of the parallel versions will also have document- and word-level readability annotations. Moreover, all three versions of each document will have the same number of paragraphs and sentences by design.

\begin{figure*}[t!]
    \centering
    \includegraphics[]{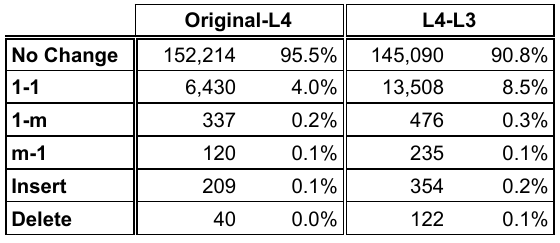}
    \captionof{table}{Lexical simplification transformation statistics in terms of replacements, insertions (Insert), and deletions (Delete) when simplifying the original text to Level~4 (Original-L4) and Level~4 text to Level~3 (L4-L3). No Change indicates no transformations. 1-1 (one-to-one), 1-m (one-to-many), and m-1 (many-to-one) indicate the different types of replacements. The percentages are calculated against the source, i.e., Original for Original-L4, and L4 for L4-L3.}\vspace{0.6cm}
    \label{tab:transformation-stats}
\end{figure*}

\section{Corpus Overview and Statistics}
\label{overview} 
\subsection{Inter-Annotator Agreement} To validate the quality of the annotations, we selected $\sim$1300 words from each book (17 paragraphs or $\sim$20K words in total) to be double annotated. Quantitatively, the differences between the annotators' texts are on par with the differences from the original text. As such 
our inter-annotator agreement check is mostly qualitative. We iterate over the doubled-annotated files word by word to investigate all differences. The double-annotated data had $\sim$6.8\% word-level mismatches when simplifying the original text to Level~4. The number of mismatches increased to 13.2\% for Level~4 to~Level~3 simplification. Most mismatches came from the fact that the annotators made different lexical simplification choices throughout the annotation process. Annotation mistakes were very infrequent and constituted $\sim$10\% of all mismatches.

\subsection{Corpus Readability Statistics} 

Table~\ref{tab:level-stats} presents the readability levels of the words in our corpus over the Original, Level~4, and Level~3 texts. After the annotation, the 15 original texts (159,265 words) resulted in 159,677 Level~4 words (0.3\% increase from the original) and 160,149 Level~3 (0.3\% increase from Level~4). Although the three versions of the texts are almost identical in size in terms of the number of words, the distribution of the readability levels of the words varies significantly.
Comparing the readability levels of the words in the original text against the ones in Level~4 (L4), there is an overall shift from higher to lower levels in  8.8\% of all original words. Specifically, all of the unknown words in the original text that are not present in the lemma-based readability lexicon (L6) were manually assigned a readability level and further simplified in case they were of Level~5 (L5). Furthermore, \textbf{all} of the Level~5 (L5) words and \textbf{some} of the Level~4 (L4) words in the original text were simplified to lower levels. This highlights that the simplification process of the original text to Level~4 involved a mix of manual corrections to the automatically assigned readability labels and lexical simplification of words that have a readability level higher than Level~4 (L4). When it comes to simplifying Level~4 (L4) text to Level~3 (L3), there is a comparable  8.3\%
change coming from simplifying Level~4 (L4) words to lower levels. Lastly, when comparing the readability levels of the words in the original text to the ones in Level~3 (L3), there is a 17.1\% decrease in the overall readability levels. 

\subsection{Transformations Statistics} 

To obtain the different types of lexical transformations that were applied throughout the annotation process, we use and extend the character edit distance (CED) word alignment tool developed by \newcite{Khalifa:2021:Character}. We obtain the word-level alignments between the original text and its Level~4 simplified version (Original-L4) and between the Level~4 text and its Level~3 simplified version (L4-L3). Each of these two alignments produces a sequence of word-level edit operations representing the lexical transformations in terms of insertions, deletions, and replacements that were a result of the manual simplification process. Each of these operations could involve more than one word at a time. 
Table~\ref{tab:transformation-stats} presents the statistics of the different lexical transformations in our corpus. The majority of the words did not involve any transformations (No Change) when simplifying the original text to Level~4 and the Level~4 text to Level~3. We quantify three types of replacements: one-to-one (1-1), one-to-many (1-m), and many-to-one (m-1) based on the number of words involved in each transformation. Out of the three replacement types, 1-1 replacements are the majority affecting 4\% of the words in the original text when it is simplified to Level~4 and 8.5\% of the Level~4 text when it is simplified to Level~3. We note that many-to-many (m-m) replacements do not occur in the corpus. When it comes to insertions and deletions, they represent small percentages of all transformations. 

We note that the difference in the number of changes between lexical transformations (Table~\ref{tab:transformation-stats}) and readability level shifts (Table~\ref{tab:level-stats}) is due to corrections in leveling without any word change, e.g.,  incorrectly labelled proper nouns are mapped to Level~0 without changing them.
\vspace{0.9cm}

\begin{figure*}[t!]
    \centering
    \includegraphics[width=\textwidth]{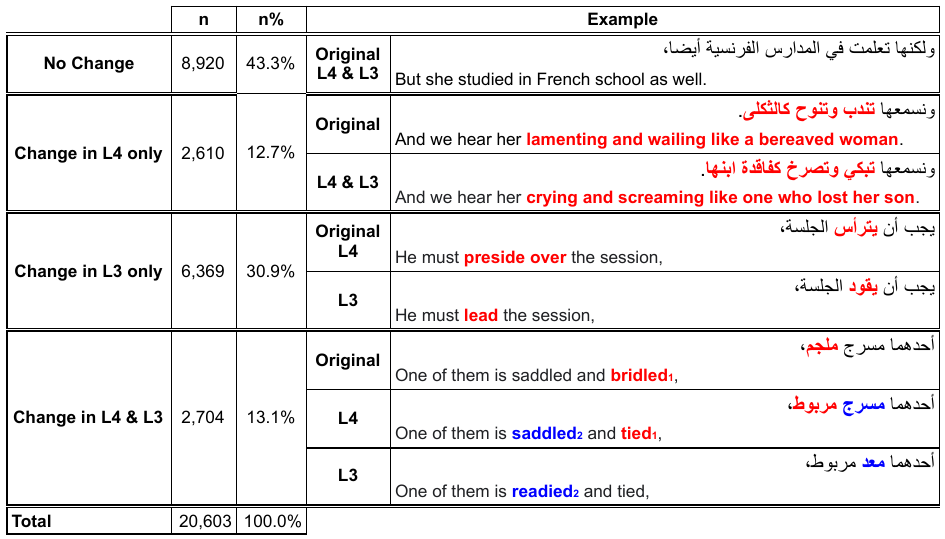}
    \captionof{table}{Statistics of fragments in our corpus based on the changes that are made to the text when it is simplified to Level~4 (L4) and then to Level~3 (L3). \textbf{n} is the number of fragments. The words in \textcolor{red}{\textbf{red}} are all of readability level 5 and the words in \textcolor{blue}{\textbf{blue}} are all of readability level 4. L4 is the simplified version of the original sentence where all level 5 words (in \textcolor{red}{\textbf{red}}) are simplified to level 4 or lower. L3 is the simplified version of the L4 sentence where all level 4 words (in \textcolor{blue}{\textbf{blue}}) are simplified to level 3 or lower.
   }
    \label{tab:fragments-stats}
\end{figure*}

\subsection{Fragments Statistics} 

After the annotation, we segmented the original paragraphs and their annotated parallels into smaller fragments based on punctuation. Since punctuation did not change throughout the annotation process, all of the parallel texts in our corpus will have the same number of punctuated fragments, and all of the fragments will be perfectly aligned. The segmentation was done carefully to guarantee that the fragments included a combination of both words and punctuation, rather than consisting solely of punctuation. We refer to the output of this segmentation process as fragments rather than sentences. This is because Arabic sentence segmentation is challenging due to the dearth of punctuation marks and the dual use of the Arabic comma (\<،>) for phrase and clause boundaries \cite{habash-etal-2022-camel}. As such these fragments may be full sentences, subordinated clauses or phrases.  
Table~\ref{tab:fragments-stats} presents the statistics of the fragments in our corpus, with associated examples. In total, there are 20,603 fragments across all texts (Original, Level~4, and Level~3). On average, each fragment consisted of $\sim$7.5 words. Overall, 43.3\% of all fragments did not have any changes, whereas 12.7\% of the fragments included changes only when the original text was simplified to Level~4 (Change in L4 only). 30.9\% of fragments included changes only when the Level~4 text was simplified to Level~3 (Change in L3 only). Lastly, 13.1\% of all fragments included changes in both simplified Level~4 and Level~3 texts. 


\subsection{Corpus Splits} To aid reproducibility when using our corpus for various research experiments, we provide train (Train), development (Dev), and test (Test) splits. We split each novel based on the number of words into Train (70\%), Dev (15\%), and Test (15\%), while respecting the full chapter boundaries. We follow the recommendations of \newcite{Diab:2013:ldc} and select full chapters from each novel such that the chapters that are in the Dev and Test sets are taken from well-separated regions of the novel. This ensures that results derived from the Dev and Test sets are not due to mere proximity of subject matter. Altogether, we end up with 113,476 words (71\%) for Train, 22,280 words (14\%) for Dev, and 23,509 words (15\%) for Test.

\section{Conclusions and Future Work}
We presented the first manually annotated Arabic parallel corpus for text simplification targeting school-aged learners. Our corpus comprises texts of 159K words selected from 15 publicly available Arabic fiction novels. Our corpus includes readability level annotations at both the document and word levels, as well as two simplified parallel versions for each text targeting learners at two different readability levels. We described the corpus selection process and outlined the guidelines we followed to create the annotations and ensure their quality.
The corpus, its parallel versions and splits, as well as, the annotation guidelines are publicly available on the SAMER Project website:\\ 
\url{http://samer.camel-lab.com/}.

In future work, we plan to extend our corpus to include text from other genres and domains. We also plan to use it in developing models for readability assessment and automatic simplification for Arabic. By building our corpus and making it publicly available, we hope to encourage research on Arabic text simplification and automatic readability assessment, as well as development of personalized Arabic pedagogical applications.


\section*{Ethics Statement}
All of the text we selected are in the public domain. All of the annotators were paid fair wages for the tasks of simplification. This effort did not require the use of extensive and compute-heavy learning models.
We acknowledge that our resource, like many others in NLP, can be used to guide controlled text generation for unethical purposes it was not intended for such as plagiaristic rewriting.

\section*{Acknowledgements}
This project was funded by a New York University Abu Dhabi Research Enhancement Fund grant. \\

\section{Bibliographical References}\label{sec:reference}

\bibliographystyle{lrec-coling2024-natbib}
\bibliography{anthology,camel-bib-v3,extra}

\section{Language Resource References}
\label{lr:ref}
\bibliographystylelanguageresource{lrec-coling2024-natbib}
\bibliographylanguageresource{languageresource,anthology,camel-bib-v3,extra}




\end{document}